\documentclass[letterpaper, 10 pt, conference]{ieeeconf}
\IEEEoverridecommandlockouts 
\usepackage{graphics} 
\usepackage{amsmath} 
\usepackage{amssymb}  
\usepackage{cite}
\usepackage{amsfonts}
\usepackage{bm}
\usepackage{breqn}

\usepackage{diagbox}
\usepackage{multirow}
\usepackage{multicol}
\usepackage{graphicx}
\usepackage[subtle,tracking=normal]{savetrees}

\usepackage{balance}
\usepackage{caption}
\usepackage{subcaption}
\usepackage{makecell}

\usepackage{hyperref}
\usepackage[percent]{overpic}
\usepackage{xcolor}
\newcommand{\insertYoutubeLink}{\url{https://youtu.be/SvfVNQ90k_w}}

\title{\LARGE \bf
Quadruped-Frog: Rapid Online Optimization \\
of Continuous Quadruped Jumping
}

\author{Guillaume Bellegarda, Milad Shafiee, Merih Ekin Özberk, Auke Ijspeert%
\thanks{
This research is supported by the Swiss National Science Foundation
(SNSF) as part of project No.197237. The authors are with the BioRobotics Laboratory, Ecole Polytechnique Federale de Lausanne (EPFL).
 {\tt \{firstname.lastname\}@epfl.ch}}
}

\begin{document}
\bstctlcite{MyBSTcontrol}
\makeatletter
\makeatother
\maketitle
\begin{abstract}
Legged robots are becoming increasingly agile in exhibiting dynamic behaviors such as running and jumping. Usually, such behaviors are either optimized and engineered \textit{offline} (i.e. the behavior is designed for \textit{before} it is needed), either through model-based trajectory optimization, or through deep learning-based methods involving millions of timesteps of simulation interactions. Notably, such offline-designed locomotion controllers cannot perfectly model the true dynamics of the system, such as the motor dynamics. In contrast, in this paper, we consider a quadruped jumping task that we rapidly optimize \textit{online}. We design foot force profiles parameterized by only a few parameters which we optimize for directly on hardware with Bayesian Optimization. The force profiles are tracked at the joint level, and added to Cartesian PD impedance control and Virtual Model Control to stabilize the jumping motions. After optimization, which takes only a handful of jumps, we show that this control architecture is capable of diverse and omnidirectional jumps including forward, lateral, and twist (turning) jumps, even on uneven terrain, enabling the Unitree Go1 quadruped to jump 0.5 $m$ high, 0.5 $m$ forward, and jump-turn over 2 $rad$. Video results can be found at \insertYoutubeLink.
\end{abstract}

\section{Introduction}
\label{sec:introduction}

Many animals exhibit a variety of dynamic and agile locomotion and jumping motor behaviors. For example, Springboks and Thomson's gazelles use stotting (or pronking) to gracefully locomote at high speeds~\cite{fitzgibbon1988stotting}. Various hypotheses have been proposed as possible explanations for this stotting behavior, including pursuit deterrence, rapid escape, anti-ambush behavior, and predator detection~\cite{caro1986functions}. Another explanation is play, often seen in baby goats. Many animals, and especially hoofed animals, are able to locomote within minutes of birth~\cite{doi:10.1126/science.1210617,garwicz2009unifying}, and baby goats can be seen to rapidly improve and fine-tune their jumping motor behaviors within their first hours/days. This is possible through primitive patterns of neural control known to exist in vertebrates, also known as Central Pattern Generators (CPGs)~\cite{ijspeert2008}, which can produce locomotion in the absence of descending drive from higher centers. 

In robotics, dynamic motor skills such as running and jumping for legged robots have recently drawn increased interest due to advances in both hardware capabilities and control architectures. Trajectory Optimization (TO) approaches have shown that optimizing over the full system dynamics allows for the generation and tracking of highly dynamic jumping motions on hardware \cite{nguyen2019jumping,chuongjump3D,katz2019mini,chignoli2022rapid,chignoli2021online}. Recent work extends such single jumps with Model Predictive Control (MPC) to transition between jumps, enabling continuous jumping on stepping stones~\cite{continuous_jump}. When the TO can be solved online, i.e. using simplified dynamics models to run as MPC, highly dynamic and robust locomotion skills can be realized on quadruped hardware\cite{dicarlo2018mpc,kim2019highly,sombolestan2021adaptive,bellicoso2018dynamic}.

Another approach to generating running and jumping controllers is through deep reinforcement learning. For example, fast trot-running \cite{ji2022concurrent,margolis2022rapid} and bounding \cite{bellegarda2022robust} have autonomously emerged end-to-end through learning frameworks. Advanced skills can also be learned by incorporating terrain-awareness for tasks such as climbing and jumping gaps~\cite{rudin2022advanced}, or rough terrain locomotion in the wild~\cite{miki2022learning}. As reward function design and exploration can be issues, other works use reference motions, careful hierarchical training schemes, and/or policy experience transfer to learn difficult jumping skills~\cite{Li-RSS-23,smith2023learning}.

\begin{figure}[!t]
    \vspace{0.06in}
    \centering
    \includegraphics[width=\linewidth]{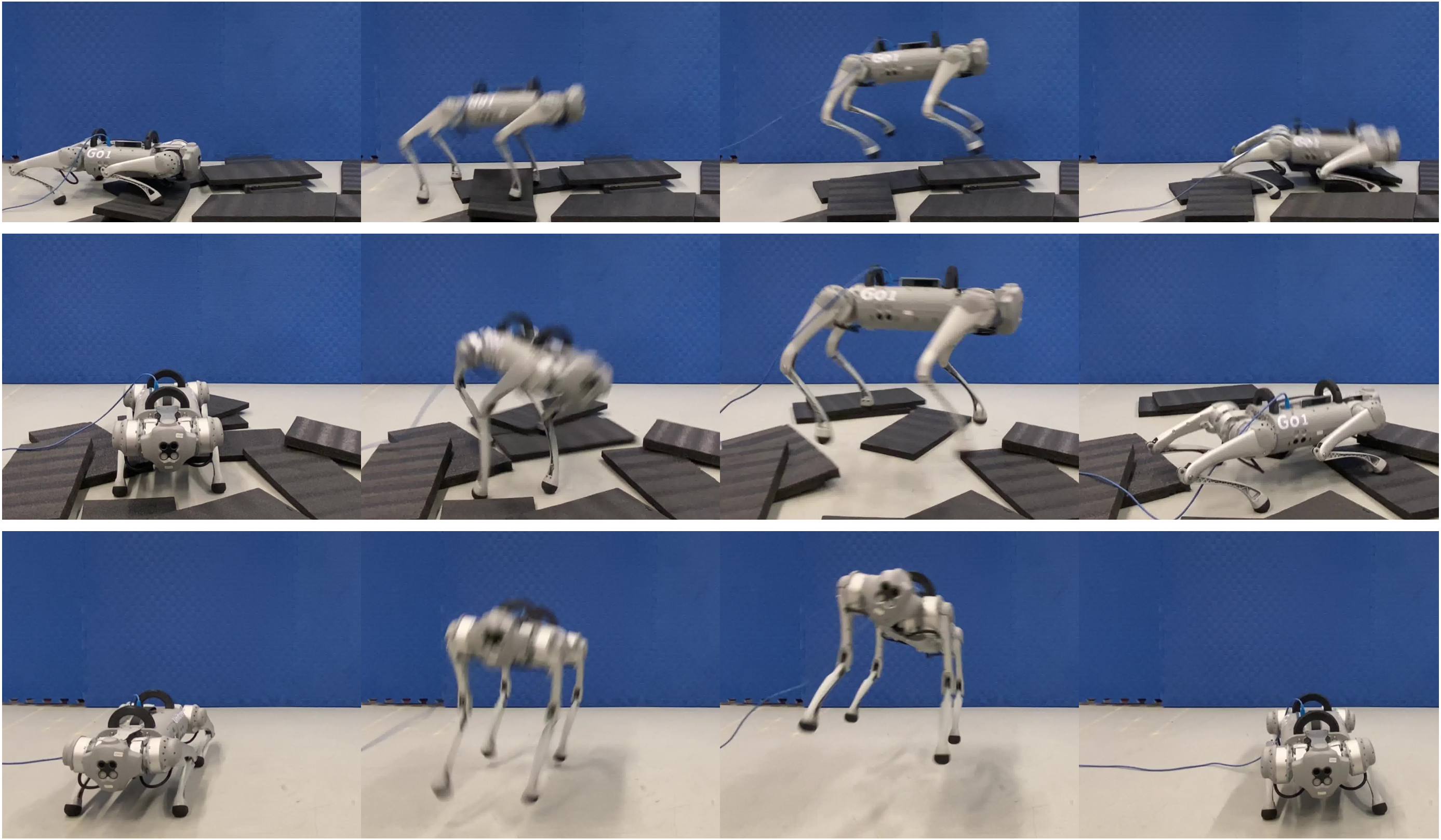} \\
    \vspace{0.1em}
    \caption{Online optimized jumping. Top: forward jumping on rough terrain (0.5 $m$ height, 0.5 $m$ distance). 
    Middle: twist jump, over 2 $rad$.
    Bottom: lateral jumping 0.3 $m$. 
    }
    \label{fig:intro}
    \vspace{-1em}
\end{figure}

Leveraging ideas and methods from both model-based control and learning-based control can help surpass the performance of either method individually~\cite{bellegarda2020online}. For example, some tasks may be difficult to model, or overly conservative, with model-based control. Conversely, learning-based methods can suffer from difficulties with appropriately exploring the state space, and reward function tuning.  Choosing an alternative action space to desired joint positions is one such beneficial combination~\cite{bellegardaIROS19TaskSpaceRL,bellegarda2022robust}. For example, learning foot position residuals on top of optimized jumping trajectories allows for jumping off of uneven terrain and with significant noise~\cite{bellegarda2020robust}. Combining MPC with reinforcement learning also allows for dynamic gap crossing capabilities~\cite{margolis2022pixels,yu2022visual,lee2022pi,yang2023cajun}, or continuous jumping through learned action residuals~\cite{pmlr-v211-yang23b}. Our previous work uses a hierarchical biology-inspired control architecture which leverages deep reinforcement learning with dynamical systems (CPGs) in the loop for both robust locomotion~\cite{bellegarda2022cpgrl,bellegarda2022visual,shafiee2023manyquadrupeds}, as well as to learn rapid gap crossing abilities~\cite{shafiee2023puppeteer,shafiee2023deeptransition}.

The above works can take significant engineering effort and time to properly tune various parameters in both optimal control frameworks (i.e.~contact timing, cost function, constraints, gains) and learning-based frameworks (i.e.~reward function design, training time, sim-to-real gap). There is also usually no online adaptation to continue to improve the methods, with some exceptions, for example continuing the training process from simulation to fine-tune locomotion controllers on hardware~\cite{smith2022legged}. This is still in contrast to the rapid adaptation of animals, which start with innate locomotion skills, and rapidly improve them with few real-world interactions. 

In this light, Bayesian Optimization has previously been applied directly on hardware to rapidly tune parameters to improve legged robot gaits for both bipeds\cite{rai2019using,rai2018bayesian,calandra2014experimental,calandra2016bayesian} and quadrupeds\cite{ruppert2022learning,widmer2023tuning}. However, such approaches have not yet been demonstrated for more dynamic locomotion skills, such as continuous jumping. 

\subsection{Contribution}
Inspired by animals and recent robotics works, we present an omnidirectional jumping controller which can be optimized online directly on hardware. We parameterize the jumps through desired force profiles to be applied at the feet, which are similar to force profiles for jumping and landing in animals such as frogs~\cite{nauwelaerts2006take}. The force profiles are tracked at the joint level, along with Cartesian PD impedance control which regulates nominal foot positions. To stabilize the jumping motions, we add Virtual Model Control. Our framework allows us to rapidly optimize dynamic jumping behaviors to continuously perform forward, lateral, and twist jumps, even under disturbances of uneven terrain and varying coefficients of friction. This online optimization is in contrast to other works which optimize jumping offline, and this allows us to optimize directly on the real system without any dynamics uncertainties or mismatches (i.e.~as is the case with simulating motor dynamics, friction, etc.). 

The rest of this paper is organized as follows. In Section~\ref{sec:method} we present our jumping parameterization design choices and integration of Bayesian Optimization. In Section~\ref{sec:result} we discuss results and analysis from optimizing our controller to perform omnidirectional and continuous jumps, even when subjected to disturbances and noise in the form of uneven terrain. Section~\ref{sec:conclusion} concludes the paper and suggests future directions for further work.

\section{Method}
\label{sec:method}

\begin{figure*}[!t]
      \centering
      \vspace{.06in}
      \includegraphics[width=\linewidth]{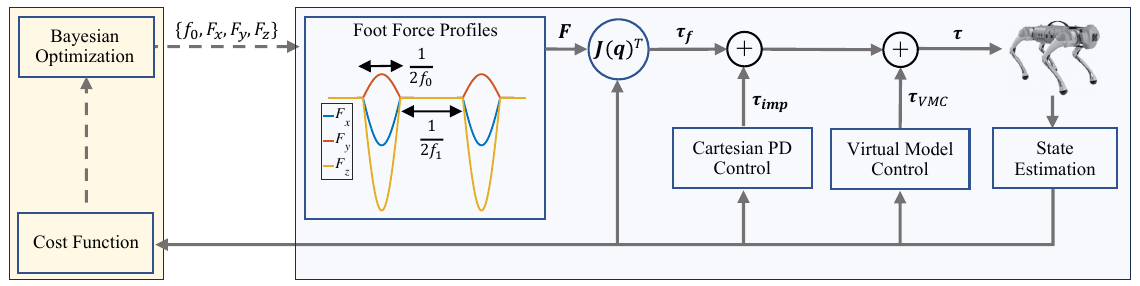}\\
      \caption{Control architecture for online jumping optimization. The right (blue) box represents the environment, where the solid arrows operate at 1 kHz. Desired foot force profiles are mapped to torques with the Jacobian. Cartesian PD impedance control helps to regulate the foot at a nominal position below the hips, and Virtual Model Control is added to help stabilize the robot and allow jumping in uneven terrain. The left (yellow) box represents the Bayesian Optimization, which selects new force profile parameters after each jump based on the accumulated cost function specifying the task. }
      \label{fig:control_diagram}
      \vspace{-1em}
\end{figure*}

In this section we describe our online jumping optimization framework and design decisions for developing omnidirectional jumping controllers for quadruped robots. Based on different cost functions associated with jumping different directions, the optimization updates the desired force profile impulses to be applied at each of the feet, varying both the frequency and magnitude of these parameters. These force profiles are tracked at the joint level with the foot Jacobian, and Cartesian PD impedance control helps to keep the feet at a nominal position beneath the hips. To avoid large Cartesian PD gains as well as improve the robot stability, we add Virtual Model Control to regulate the base roll and pitch. This additionally allows the robot to jump in uneven terrains. A high-level control diagram is illustrated in Figure~\ref{fig:control_diagram}, and we explain all components below.

\subsection{Generating Jumping Motion Behaviors}

Jumping is a dynamic and coordinated movement that relies on the intricate interplay of muscles and neural circuits. At the heart of this impressive skill are Central Pattern Generators (CPGs), specialized neural networks found within the spinal cord that generate rhythmic motor patterns. These neural circuits play a crucial role in orchestrating the precise sequence of muscle contractions required for a successful jump. By producing the necessary motor commands and coordinating the timings of muscle activations, CPGs enable organisms, from frogs to humans, to execute powerful and well-timed leaps, making jumping an intriguing example of the neural control of complex movements.

To represent the CPG circuits in the spinal cord for generating quadruped locomotion skills, a number of abstract oscillators have been proposed, with some of the most popular including Matsuoka oscillators~\cite{matsuoka1987mechanisms} and phase oscillators~\cite{righetti08,ijspeert2007salamander}. These typically generate rhythm in either joint space~\cite{ijspeert2007salamander,sprowitz2013cheetah} or task space (i.e. for each limb)~\cite{righetti08,bellegarda2022cpgrl}, with coupling between different joints and limbs to produce different gaits. In our previous works~\cite{shafiee2023puppeteer,shafiee2023deeptransition}, we used deep reinforcement learning to modulate the CPG, and thus task space positions, in order to dynamically locomote across gaps. 

\begin{figure}
    \centering
    \includegraphics[width=\linewidth]{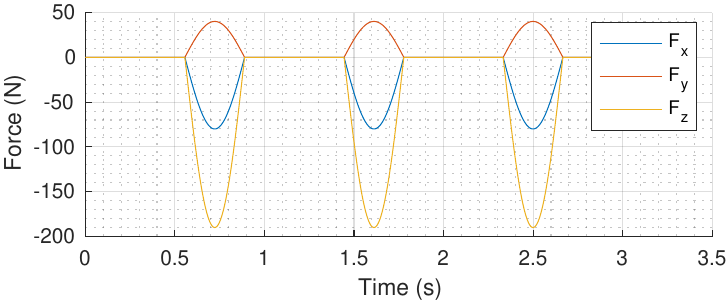}
    \vspace{-1em}
    \caption{Force trajectories for hopping forwards and right in the body frame. When the impulse is not active, the system is in the air, or landing. Parameter $f_0$ determines the frequency of the impulse, and parameter $f_1$ is the frequency between impulses (and is not optimized). 
    }
    \label{fig:bo_traj}
    \vspace{-0.5em}
\end{figure}

\begin{figure}[!tpb]
    \centering
    \includegraphics[width=\linewidth]{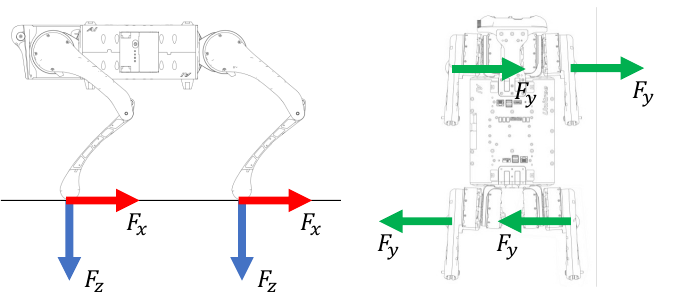}
    \vspace{-1em}
    \caption{
    Jumping force directions visualized at the feet of the Unitree Go1 quadruped. Left: planar $XZ$ forces applied at the feet for jumping forward. Right: top view of lateral forces for performing a counterclockwise twist jump.  
    }
    \label{fig:bo_forces}
    \vspace{-0.7em}
\end{figure}

In contrast, in this paper, due to the inherent force interaction during jumping, we choose to optimize force profiles rather than joint or task space positions. Notably, this essentially re-formulates the abstract oscillators that are typically used to model the CPGs into force space. We take inspiration from frogs which have approximate half-sine wave force profiles for both take-off and landing in both horizontal and vertical directions~\cite{nauwelaerts2006take}. 

As in previous works using abstract oscillators, we monitor the system state with the (uncoupled) phase oscillator:
\begin{align}
    \dot{\theta} &= 2 \pi f_i \ , \quad \text{where} \ f_i =
 \begin{cases}
    f_0 & \text{if } \pi \leq \theta < 2\pi  \\
    f_1 & \text{if }  0 \leq \theta < \pi
\end{cases} 
\end{align}
where $\theta$ is the phase of the oscillator. The frequency $f_i$ changes based on the phase, which dictates whether to apply the impulse force, or to be inactive (for example during flight, or time between jumps). We define the impulse phase frequency as $f_0$ (to apply the force with the foot in contact with the ground) as $\pi \leq \theta < 2\pi$, and the off-phase frequency as $f_1$ when $0 \leq \theta < \pi$. 
Therefore, the impulse time duration $T_{\mathrm{impulse}}$ will be $\frac{1}{2f_0}$, and the time between impulses will be $T_{\mathrm{off}} = \frac{1}{2f_1}$. 
Given these phase relationships, Figure~\ref{fig:bo_traj} shows the force profiles and directions for a set of parameters for setting the 3D force vector at the foot of each leg:
\begin{align}
\bm{F}_i =
 \begin{cases}
    [F_x \  F_y \ F_z]^\top \sin(\theta) & \text{if } \sin(\theta) < 0 \\
    \bm{0}_{3 \times 1} & \text{otherwise}
\end{cases} 
\end{align}

\noindent where $F_x$, $F_y$, $F_z$ are the force amplitudes applied at the foot contact in the local frame. By coordinating the signs of the forces $F_x$ and $F_y$, jumping can be accomplished in different directions, for example jumping forward by applying forces in the $X$ and $Z$ directions for all feet (Figure \ref{fig:bo_forces} left), or a twist turn by changing the direction of the applied $Y$ force for the front and rear feet (Figure \ref{fig:bo_forces} right).

\subsection{Leg Controller}
The forces from the above force profiles can be applied at each foot by mapping the forces to torques at the joint level with:  
\begin{align}
    \bm{\tau}_{\mathrm{f}} = \bm{J}(\bm{q})^\top \bm{F}
\end{align}
where $\bm{J}(\bm{q})$ is the foot Jacobian at joint configuration $\bm{q}$. We also add a Cartesian PD impedance controller to regulate the foot to a nominal position below the hips. The foot position error is mapped to torques and tracked at the joint level with the following controller for each leg $i$: 
\begin{align}
    \bm{\tau}_{\mathrm{imp}} &= \bm{J}(\bm{q})^\top \Bigl[ \bm{K}_{p} \left(\bm{p}_{d} - \bm{p} \right) - \bm{K}_{d} \left( \bm{v} \right)  \Bigr] - \bm{K}_{d,joint} (\bm{\dot{q}})
    \label{eqn:leg_ff}
\end{align}
where $\bm{K}_p$ and $\bm{K}_d$ are diagonal matrices of proportional and derivative gains in Cartesian coordinates to track the desired foot positions $(\bm{p}_d)$ with zero desired foot velocity $(\bm{v})$ in the leg frame. We add a small joint damping term for stability in the hardware experiments. We use $\bm{K}_{p}=400\bm{I}_3,\ \bm{K}_{d}=8\bm{I}_3, \  \bm{K}_{d,joint}=0.8\bm{I}_3$. 

\subsection{Virtual Model Control}
\label{sec:vmc}
To improve the stability of the system, robustness to uneven terrain, and avoid large Cartesian space gains, we add Virtual Model Control (VMC) while in contact with the ground, similar to~\cite{mos2013cat,mos2013oncilla}. We attach virtual springs between a hypothetical XY plane through the center of the trunk, and with another plane horizontal with respect to the world coordinates. These virtual springs naturally generate forces to adjust the attitude (pitch and roll) of the body to be parallel to the ground: 
\begin{align}
    \bm{P} = \bm{R} \begin{bmatrix}
 1 & 1 & -1 & -1\\
-1 & 1 & -1 &  1 \\
 0 & 0 &  0 &  0
\end{bmatrix} \\
\bm{F}_{VMC} = \begin{bmatrix} 
\bm{0}_{2 \times 4} \\
k_{att} ( \begin{bmatrix}
0 & 0 & 1
\end{bmatrix} \bm{P}) 
\end{bmatrix}
\end{align}

\noindent where $\bm{R}$ is the robot's rotation matrix with respect to the world coordinates, $\bm{P}$ are the relative coordinates of the corners of the virtual plane through the body, $k_{att}=200$ is the gain, and the columns of $\bm{F}_{VMC}$ are the virtual forces to be added to each leg $i$ (Front Right, Front Left, Rear Right, Rear Left). 

The VMC force contribution column $i$ is added to the feedforward force controller and impedance controller for leg $i$ with:
\begin{align}
    \bm{\tau}_{\mathrm{VMC},i} = \bm{J}_i(\bm{q}_i)^\top \bm{F}_{VMC_i}
\end{align}
making the full torque vector for each leg $i$: 
\begin{align}
    \bm{\tau}_i =  \bm{\tau}_{\mathrm{f},i} + \bm{\tau}_{\mathrm{imp},i} + \bm{\tau}_{\mathrm{VMC},i} 
\end{align}

\subsection{Bayesian Optimization}
\label{sec:bo}

Bayesian optimization is a powerful and versatile approach to solving complex optimization problems. It is a probabilistic model-based optimization technique that leverages Bayesian statistics to efficiently explore and exploit the parameter space of a function to find the optimal solution. Unlike more traditional optimization methods which generate jumping maneuvers offline~\cite{nguyen2019jumping,chuongjump3D}, Bayesian optimization is particularly well-suited for online problems with noisy or expensive-to-evaluate objective functions, where it strives to strike a balance between exploration (searching for promising areas) and exploitation (focusing on areas likely to yield the best results). 

In Bayesian optimization, a probabilistic model (typically Gaussian Process or Tree-Parzen Estimator (TPE)) is built to represent the unknown objective function. TPE uses a tree-structured approach to model the objective function as a combination of two probability distributions: one for the promising parameter configurations, and another for the less promising ones. These distributions guide the exploration and exploitation of the parameter space. TPE adaptively selects and evaluates candidate configurations that are more likely to improve the objective, which makes it highly sample-efficient. In this paper, we use the TPE implementation of Optuna~\cite{optuna_2019} to optimize several omnidirectional jumping tasks. 

\subsubsection{Forward Jumping} 
The first task is continuous jumping forward. The cost function is defined as the difference between the final $x$ position in the world frame after the jump, and the initial $x$ position before the jump:
\begin{align}
    J_{\mathrm{fwd}} = x_{\mathrm{final}} - x_{\mathrm{init}}
\end{align}

\subsubsection{Lateral Jumping} 
The second task is continuous lateral jumping. The cost function is defined as the difference between the final $y$ position in the world frame, and the initial $y$ position. The sign can be changed  for jumping left ($+$) in the body frame, or ($-$) for jumping right:
\begin{align}
    J_{\mathrm{lat}} = (\pm) \  ( y_{\mathrm{final}} - y_{\mathrm{init}} ) 
\end{align}

\subsubsection{Twist Jumping} 
The third task is continuous twist jumping (yaw rotational jump). The cost function is defined as the difference between the the final yaw angle and initial yaw angle in the world frame. The sign can be changed for twist jumping counterclockwise ($+$) in the body frame, or ($-$) for jumping clockwise:
\begin{align}
    J_{\mathrm{twist}} = (\pm) \  ( \psi_{\mathrm{final}} - \psi_{\mathrm{init}} )
\end{align}

In case the robot falls during a jump due to a poor selection of parameters, we set $J=0$ for that iteration of the optimization. This is to avoid exploiting the dynamics to get a high objective value even though the robot may no longer be standing.

\section{Experimental Results and Discussion}
\label{sec:result}

In this section we report and discuss results from the online optimization of our jumping controllers. Sample snapshots of the quadruped jumping in uneven terrain are shown in Figure~\ref{fig:intro}, and the reader is encouraged to watch the supplementary video for clear visualizations of the discussed experiments. 

\subsubsection{Implementation Details} We use the Unitree Go1 quadruped~\cite{unitreeGO1}, and the Optuna library implementation of the TPE algorithm for the Bayesian optimization~\cite{optuna_2019}. Initial tests and optimizations are carried out in simulation in Gazebo before moving to hardware. The parameters and their ranges we optimize are listed in Table~\ref{tab:bo_ranges}. The Cartesian PD and Virtual Model Control gains are tuned once on flat terrain so the robot can maintain balance and remain standing. We compute the cost function and run the Bayesian optimization to update the parameters after each jump for single jumps. Each iteration thus corresponds to a single jump. 

\begin{table}[tpb]
\centering
\vspace{0.06in}
\caption{Optimization parameter ranges for generating forward, lateral, and twist (turn) jumping.
}
\vspace{-0.4em}
\begin{tabular}{ c c c c  }
Parameter & Lower Bound & Upper Bound & Units \\
\hline
$f_0$ & 0.75 & 1.75 & Hz \\
$F_x$ & 0 & 150 & N \\
$F_y$ & 0 & 150 & N \\
$F_z$ & 150 & 350 & N \\
\hline
\end{tabular} \\
\label{tab:bo_ranges}
\end{table}

\begin{figure}[t]
    \vspace{0.06in}
    \centering
    \includegraphics[width=0.95\linewidth]{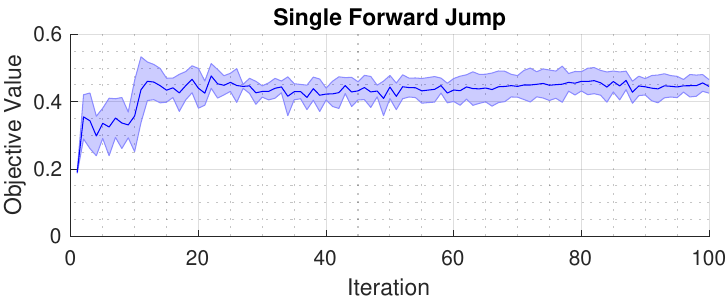}\\
    \vspace{0.2em}
    \includegraphics[width=0.95\linewidth]{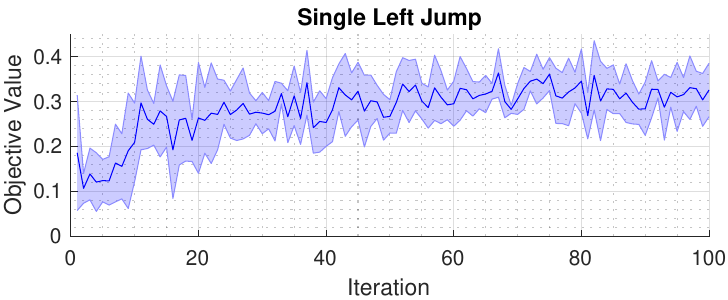}\\
    \vspace{0.2em}
    \includegraphics[width=0.95\linewidth]{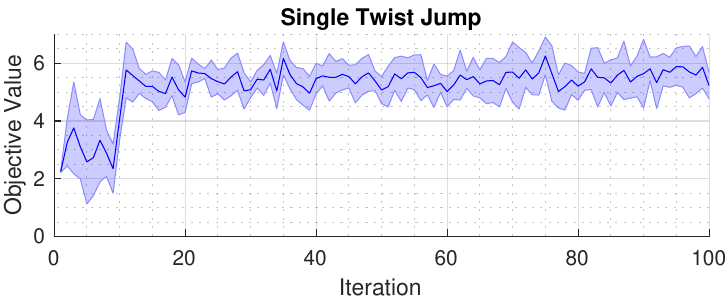}\\
    \caption{
    Training curves averaged across 5 runs with different random seeds for optimizing forward (top), lateral left (middle), and twist-turn  (bottom) jumps in Gazebo. All runs result in successful jumping. 
    }
    \label{fig:bo_training}
    \vspace{-0.8em}
\end{figure}

\begin{figure}[!tpb]
    \centering
    \includegraphics[width=\linewidth]{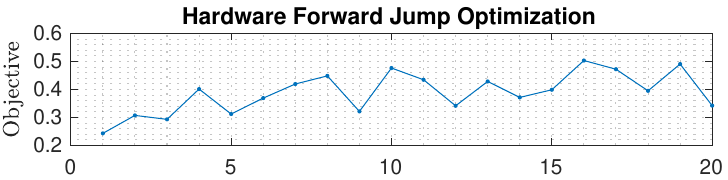}\\
    \includegraphics[width=\linewidth]{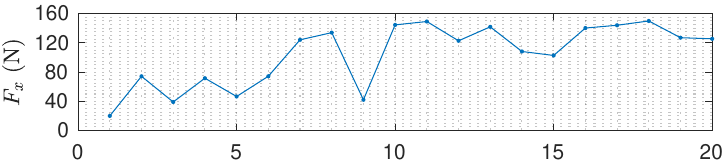}\\
    \includegraphics[width=\linewidth]{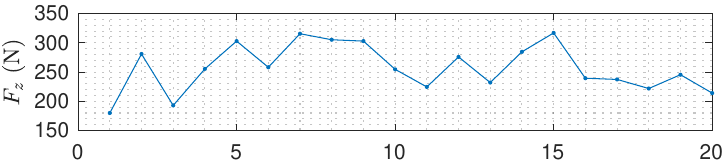}\\
    \includegraphics[width=\linewidth]{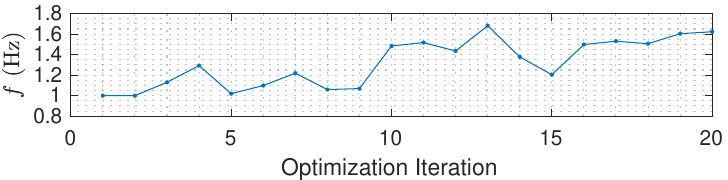}\\
    \caption{
    Forward jumping optimization on the Unitree Go1 hardware in 20 iterations. From top to bottom: objective value (forward distance jumped), force profile parameter $F_x$, force profile parameter $F_z$, and frequency $f$. Notably, there is a trade-off between forward/downward force that must occur when jumping forward. 
    }
    \label{fig:bo_fwd_hw}
    \vspace{-1em}
\end{figure}

\begin{figure}[!tpb]
    \vspace{0.06in}
    \centering
    \includegraphics[width=\linewidth]{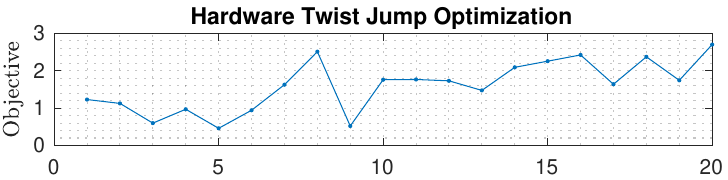}\\
    \includegraphics[width=\linewidth]{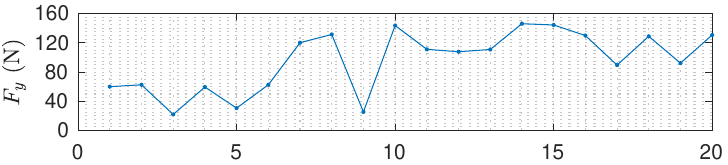}\\
    \includegraphics[width=\linewidth]{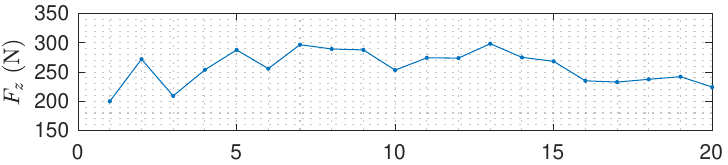}\\
    \includegraphics[width=\linewidth]{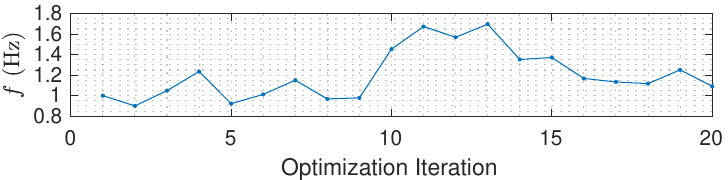}\\
    \caption{
    Twist-turn jumping optimization on the Unitree Go1 hardware in 20 iterations. From top to bottom: objective value (yaw rotation jumped), force profile parameter $F_y$, force profile parameter $F_z$, and frequency $f$. Notably, there is a trade-off between lateral/downward force that must occur when twist-jumping to both jump high enough, as well as generate a large moment. 
    }
    \label{fig:bo_twist_hw}
    \vspace{-0.9em}
\end{figure}

\subsubsection{Optimization Convergence} 
\begin{figure*}[!t]
    \vspace{0.06in}
    \centering
    \includegraphics[width=\linewidth]{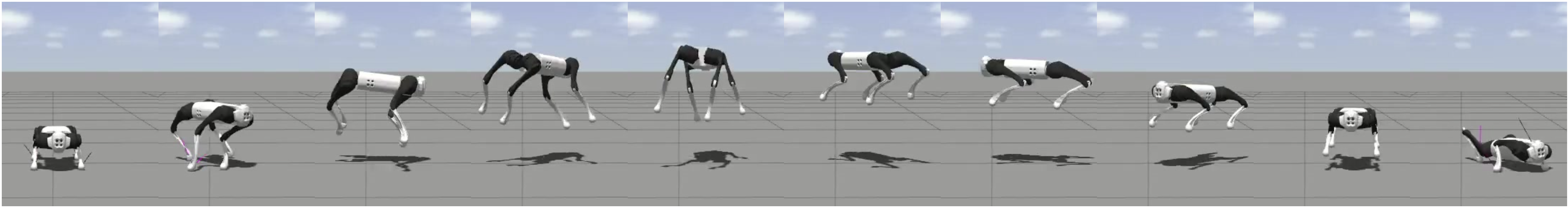}\\
    \vspace{-0.2em}
    \caption{Optimized twist-turn jump in Gazebo. The Unitree Go1 quadruped jumps 0.67 m high while rotating over 360 degrees. }
    \label{fig:twist_gazebo}
    \vspace{-0.7em}
\end{figure*}

We first investigate the sample-efficiency and optimization time of applying Bayesian optimization to tune the force profiles for our jumping control architecture. For each type of jump, we run 5 different optimizations with different random seeds and observe the robot's ability to successfully complete the desired jumps. 
Figure~\ref{fig:bo_training} shows the mean objective value across the 5 random seeds for each of the three jumping tasks vs. Bayesian Optimization iteration in the Gazebo simulation. In all cases, we can observe the rapid convergence and improvement of the quadruped's ability to jump forward, laterally, and twist-turn. Due to the ideal modeling (i.e. motors are not modeled, the feet do not slip as the coefficient of friction is high), the robot can typically jump farther and better in simulation. For example, the robot is able to optimize a twist-turn jump of 0.67 $m$ in height and over 360 degrees in rotation in simulation while respecting the torque limits of the robot, as shown in Figure~\ref{fig:twist_gazebo}. 

\begin{figure}[!t]
    \centering
    \includegraphics[width=\linewidth]{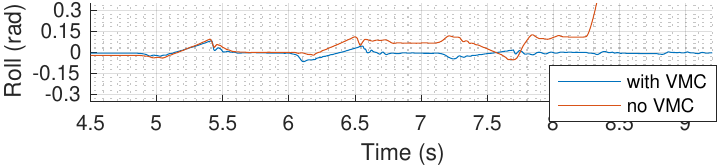}
    \includegraphics[width=\linewidth]{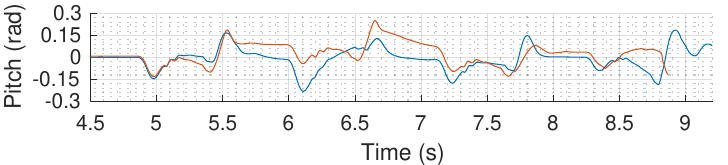}
    \includegraphics[width=\linewidth]{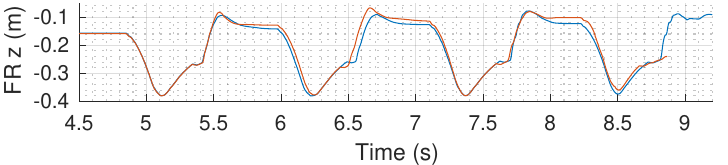}
    \includegraphics[width=\linewidth]{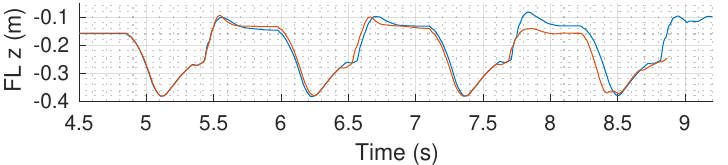}
    \includegraphics[width=\linewidth]{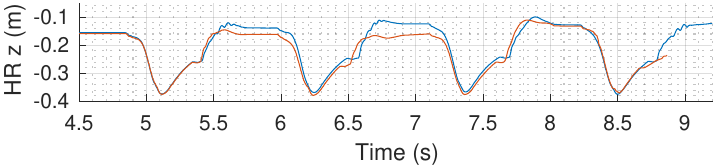}
    \includegraphics[width=\linewidth]{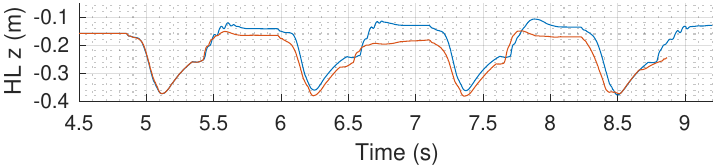}
    \caption{Rough terrain adaptation with and without Virtual Model Control active. From top: (1) base roll, (2) base pitch, (3)-(6) Front Right (FR), Front Left (FL), Hind Right (HR), Hind Left (HL) foot $z$ positions in the leg frame.  The jump at 8.3 (s) without the VMC causes a fall due to the initial roll angle.  }
    \label{fig:vmc}
    \vspace{-0.7em}
\end{figure}

Figure~\ref{fig:bo_fwd_hw} shows the optimization process for optimizing forward jumping directly on the hardware. We observe that the objective value increases during the training, which is the distance that the quadruped jumps forward. Similarly to in simulation, we see rapid convergence to being able to jump 0.5 $m$ forward. In order to accomplish such a jump, the optimizer selects the parameters $F_x$, $F_z$, and $f$, which are also shown in the Figure. Notably, there must be a trade-off between these parameters, which must maintain a ratio to jump high enough so that the robot does not slip and fall forward, but also not too high so that the robot does not make any forward progress. The best jumps can be seen when $F_z$ is not at its maximum. Interestingly, the best jump occurs at Trial 17, where $F_x = 140 N$ and $F_z = 239 N$, corresponding to an overall take-off angle (from the force direction) of approximately 60 degrees. This is in line with observations from biological frogs~\cite{nauwelaerts2006take}, as well as robotic frogs~\cite{fan2022design}. 

Figure~\ref{fig:bo_twist_hw} shows the optimization process for optimizing a twist-turn jump directly on the hardware. As in the forward case, the robot rapidly improves its jump-turning abilities within 20 trials. There is again the trade-off between applying the forces $F_y$ (in opposite directions for the front/rear feet to generate the moment), and the magnitude of the applied downward force $F_z$.  Due to the motor dynamics and coefficient of friction causing the feet to slip at take-off, the robot cannot rotate as much during the twist jump on the hardware compared to in simulation. 

\subsubsection{Continuous Jumping Optimization} 
By decreasing the time between successive jumps, we are able to accomplish continuous jumping. This can be done by increasing the frequency $f_1$, shown in Figure~\ref{fig:control_diagram}. However, with decreased time between jumps, the landing configuration becomes increasingly important in order for the next applied forces to be able to successfully have the same outcome between jumps. Several jumps can also be optimized together, where instead of running the optimization after each individual jump, we can let the robot jump several times and then compute optimal parameters that may work better for successive jumps rather than individual jumps. Another possibility would be adding the frequency of jumps $f_1$ to the optimization parameters. 

\subsubsection{Rough Terrain Adaptation}
We test the robustness of the continuous jumping controller by randomly placing overlapping 0.03 $m$ foam blocks and weights on the floor. The foam blocks are very light and are easily kicked and moved by the robot, causing additional disturbances due to sliding between the robot feet and the floor. Our controller is robust to such disturbances for omnidirectional jumps, though as can be expected, the noise can cause the robot to not jump as far or turn as much as in  nominal conditions. Snapshots are shown in Figure~\ref{fig:intro}, and experiments can be seen in the video. 

The Virtual Model Control block is especially important in such environments. We tested removing the VMC while jumping forward on the rough terrain, and after a few jumps, keeping the feet at their nominal heights in contact phase despite the uneven terrain caused a large initial roll angle during take-off, causing a fall (see video). This can be observed in Figure~\ref{fig:vmc}, where at 8 seconds, we see that there is non-zero roll without the VMC active (red line), and the foot positions do not compensate for this (left feet are still extended). This leads to the large roll take-off and subsequent flip and fall. In contrast, in the same scenario at 8 seconds, with the VMC active (blue line), we see that the left legs are retracted to keep the roll at zero, enabling a successful jump without falling. We could expect even better performance with a more sophisticated whole-body controller, which we plan to add in future work. 

\section{Conclusion}
\label{sec:conclusion}

In this paper, we have presented a control architecture for rapidly optimizing quadruped jumping skills directly on hardware. We designed and parameterized force profiles with only a few parameters, which can be rapidly optimized within just a few jumps. 
The force profiles are tracked at the joint level with the foot Jacobian, and Cartesian PD impedance control helps to keep the feet at a nominal position beneath the hips. We also added Virtual Model Control to help stabilize the jumping skills to allow for continuous jumping even on uneven terrain. We demonstrated omnidirectional skills including jumping forward (0.5 $m$ in height, 0.5 $m$ in distance), jumping laterally, and twist-turn jumps (over 2 $rad$). 

There are several directions for future work. While VMC helped to stabilize the jumping motions, a whole-body controller could help with even better robustness. This could be either model-based, or trained with deep reinforcement learning as a feedback control policy on top of the force profiles, similar to recent work learning residual actions on top of a base jumping policy~\cite{pmlr-v211-yang23b}. Another direction could be to integrate a landing controller to dampen the impact, for example as presented in~\cite{jeon2022online} or \cite{roscia2023reactive}, or study and combine the effects of passive or hybrid compliance components in parallel with the motors~\cite{ashtiani2021hybrid}. Lastly, it would be interesting to measure the ground reaction forces at the feet with force plates to verify and improve the application of the desired forces.  

There is also opportunity for drawing connections to biology, for example comparisons with force profiles from different animal species to improve the parametrization and design. This could also be used to map the force profiles back to muscle activations to study mechanisms of flexor and extensor profiles that could be formed by the CPG. Our framework can also be used to study questions related to front limb damping on landing as shown by the biological study on frogs~\cite{nauwelaerts2006take}, and effects of sensory feedback on jumping and landing ability (i.e.~the importance of the vestibular system (VMC or whole-body control))~\cite{cox2018influence}.


\bibliographystyle{IEEEtran}
\bibliography{refs}

\end{document}